\documentclass[a4paper]{article}
\usepackage[usenames, dvipsnames]{xcolor}
\usepackage{acl2020}
\usepackage{times} 
\usepackage{stmaryrd}
\usepackage[shortlabels]{enumitem}
\usepackage{microtype}
\usepackage[utf8x]{inputenc}
\usepackage{url}
\usepackage[hang,flushmargin]{footmisc}
\usepackage{colortbl,booktabs}
\usepackage{tikz}

\usepackage{pgfplots}

\definecolor{gold}{HTML}{fcca4c}
\definecolor{high}{HTML}{8f89c4}
\definecolor{med}{HTML}{90aade}
\definecolor{low}{HTML}{8bbcfc}
\newcolumntype{g}{>{\columncolor{gold}}c}
\newcolumntype{h}{>{\columncolor{high}}c}
\newcolumntype{m}{>{\columncolor{med}}c}
\newcolumntype{w}{>{\columncolor{low}}c}
\usepackage{mdsymbol}
\usepackage{clock}
\usepackage[clock]{ifsym}
\usepackage{multirow}
\usepackage{dblfloatfix}
\usepackage{tipa}
\everypar{\looseness=-1}
\newcommand{\dd}[1]{\textcolor{Green}{\bf\footnotesize +#1}}
\newcommand{\cld}[1]{\textcolor{Blue}{\footnotesize #1$\times$}}
\newcommand{\blue}[1]{\textcolor{RoyalBlue}{#1}}
\newcommand{\purple}[1]{\textcolor{Violet}{#1}}
\newcommand{\orange}[1]{\textcolor{Orange}{#1}}
\newcommand{\clk}[0]{\ClockFrametrue\ClockStyle=2\clock{11}{5}}

\usepackage{todonotes}
\usepackage{twoopt}
\newcommandtwoopt*{\myref}[3][][]{%
  \hyperref[{#3}]{%
    \ifx\\#1\\%
    \else
      #1~%
    \fi
    \ref*{#3}%
    \ifx\\#2\\%
    \else
      \,#2%
    \fi
  }%
}
\title{Phone Features Improve Speech Translation}

\newcommand{\jhu}{\textrm{\normalfont \textipa{7}}}
\newcommand{\cmu}{\textrm{\normalfont \textbeltl}}

\author{Elizabeth Salesky$^\jhu$~\;~ \and Alan W Black$^\cmu$\\
  $^\jhu$Johns Hopkins University \\ 
  $^\cmu$Carnegie Mellon University \\
  \texttt{esalesky@jhu.edu, awb@cs.cmu.edu}
}

\aclfinalcopy % Uncomment this line for the final submission
\def\aclpaperid{2572} %  Enter the acl Paper ID here

\begin{document}
\maketitle

\begin{abstract}

End-to-end models for speech translation (ST) more tightly couple speech recognition (ASR) and machine translation (MT) than a traditional cascade of separate ASR and MT models, with simpler model architectures and the potential for reduced error propagation. 
Their performance is often assumed to be superior, though in many conditions this is not yet the case. 
We compare cascaded and end-to-end models across high, medium, and low-resource conditions, and show that cascades remain stronger baselines. 
Further, we introduce two methods to incorporate phone features into ST models. 
We show that these features improve both architectures, closing the gap between end-to-end models and cascades, and outperforming previous academic work -- by up to 9 BLEU on our low-resource setting. 

\end{abstract}

%---------------------------------------------
\section{Introduction}

End-to-end models have become the common approach for speech translation (ST), but the performance gap between these models and a cascade of separately trained speech recognition (ASR) and machine translation (MT) remains, particularly in low-resource conditions. 
Models for low-resource ASR leverage phone\footnote{The term `phone' refers to segments corresponding to a collection of fine-grained phonetic units, but which may separate allophonic variation: see \citet{jurafsky2000speech}.} information, but this information is not typically leveraged by current sequence-to-sequence ASR or speech translation models. 
We propose two methods to incorporate phone features into current neural speech translation models. 
We explore the existing performance gap between end-to-end and cascaded models, and show that  incorporating phone features not only closes this gap, but greatly improves the performance and training efficiency of both model architectures, particularly in lower-resource conditions.

The sequences of speech features used as input for ST are $\approx$10 times longer than the equivalent sequence of characters in e.g. a text-based MT model. 
This impacts memory usage, the number of model parameters, and training time. 
Multiple consecutive feature vectors can belong to the same phone, but the exact number depends on the phone and local context. 
Further, these speech features are continuously valued rather than discrete, such that a given phone will have many different instantiations across a corpus. 
Neural models learn to associate ranges of similarly valued feature vectors in a data-driven way, impacting performance in lower-resource conditions. 
Using phoneme-level information provides explicit links about local and global similarities between speech features, allowing models to learn the task at hand more efficiently and yielding greater robustness to lower-resource conditions.  
\looseness=-1

We propose two simple heuristics to integrate phoneme-level information into neural speech translation models: (1) as a more robust intermediate representation in a cascade; and (2) as a concatenated embedding factor. 
We use the common Fisher Spanish--English dataset to compare with previous work, and simulate high-, mid-, and low-resource conditions to compare model performance across different data conditions. 
We compare to recent work using phone segmentation for end-to-end speech translation \citep{salesky2019acl}, and show that our methods outperform this model by up to 20 BLEU on our lowest-resource condition.\footnote{4-reference BLEU scores are used for this dataset.}
Further, our models outperform all previous academic work on this dataset, achieving similar performance trained on 20 hours as a baseline end-to-end model trained on the full 160 hour dataset. 
Finally, we test model robustness by varying the quality of our phone features, which may indicate which models will better generalize across differently-resourced conditions.\footnote{Our code is public:  \url{github.com/esalesky/xnmt-devel}}
\looseness=-1

%--------------------------------------------
%here to force placement at top of next page
%--------------------------------------------
\begin{figure*}[ht]
\centering
  \includegraphics[width=1\linewidth]{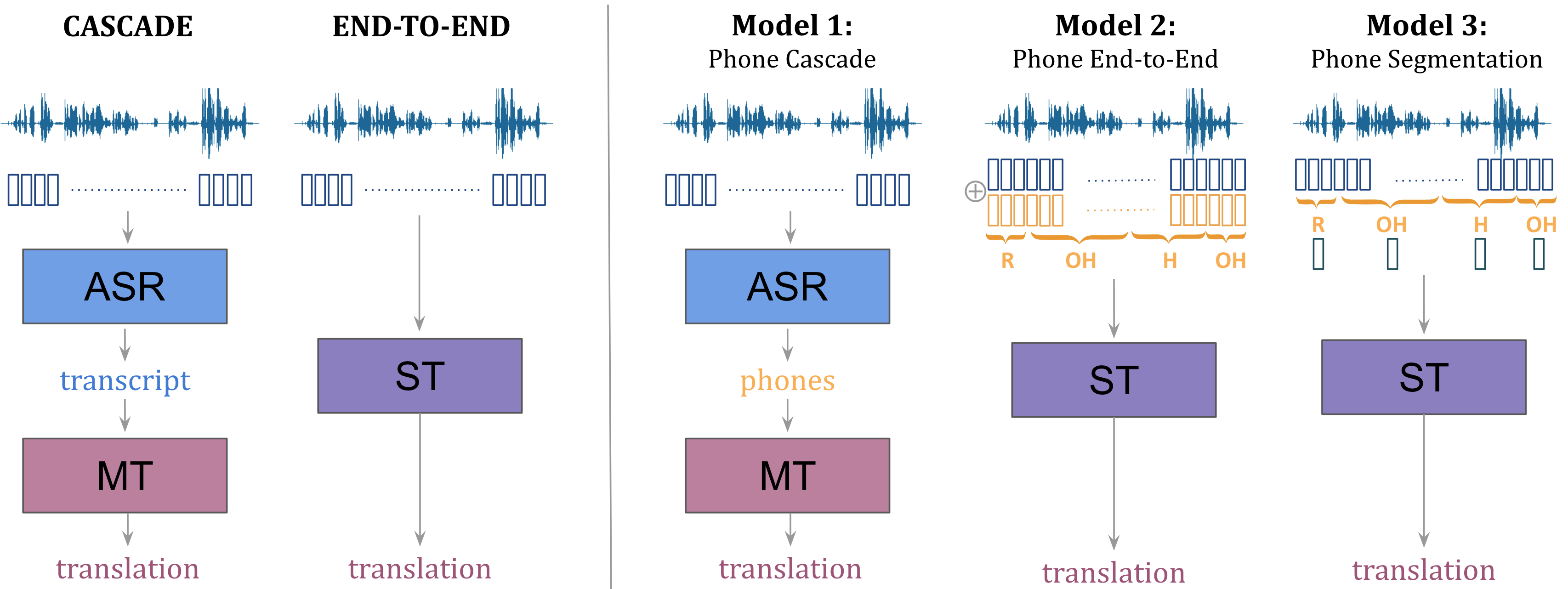}
  \caption{Comparison between traditional cascaded and end-to-end models, and our proposed methods using \textit{phone features} as (1) the intermediate representation in a cascaded model; and (2) a concatenated embedding factor in an end-to-end model. We additionally compare to previous work; (3) where phone segmentation is used for feature vector downsampling in time \citep{salesky2019acl}.}
  \label{models}
\vspace{-1em}
\end{figure*}
%------------------------------------------

\section{Models with Phone Supervision}

We add higher-level phone features to low-level speech features to improve our models' robustness across data conditions and training efficiency. 
We propose two methods to incorporate phone information into cascaded and end-to-end models, depicted in \myref[Figure]{models}. 
Our \textit{phone cascade} uses phone labels as the machine translation input, in place of the output transcription from a speech recognition model. 
Our \textit{phone end-to-end} model uses phone labels to augment source speech feature vectors in end-to-end models.
We call these end-to-end or `direct' because they utilize a single model with access to the source speech features, though they additionally use phone features generated by an external model. 
We additionally compare to a recent end-to-end model proposed by \citet{salesky2019acl}.

\paragraph{Model 1: Phone Cascade.} 
In a cascade, the intermediate representation between ASR and MT is the final output of a speech recognition model, e.g. characters, subwords, or words. 
Using separate models for ASR and MT means that errors made in ASR are likely to propagate through MT. 
Common errors include substitution of phonetically similar words, or misspellings due to irregularities in a language's orthography, the latter of which may be addressed by using phone labels in place of ASR output. 
By not committing to orthographic targets, we believe this model will propagate fewer errors to downstream MT. 

\paragraph{Model 2: Phone End-to-End.}
Our final model uses phone-factored embeddings, where trainable embeddings for phone features are concatenated to typical speech feature vector input.
Because phone durations are variable and typically span more than one filterbank feature (or frame), adjacent filterbank features may have the predicted phone label; in the example shown in \myref[Figure]{models}, /R/ spans three frames or filterbank features. 
We note that this method maintains the same source sequence length as the original speech feature sequence. 
This method associates similar feature vectors at the corpus level, because all filterbank features with the same phone alignment (e.g. /OH/) will have the same trainable phone embedding concatenated. 
In MT and NER, concatenating trainable embeddings for linguistic features to words, such as morphemes and phones, has improved models' ability to generalize \cite{sennrich2016linguistic,chaudhary2018adapting}.
While these works appended finer-grained information to associate words with similar lower-level structure, we use phone embeddings to associate higher-level structure to similar but unique speech feature vectors globally across a corpus. 

\paragraph{Model 3: Phone Segmentation.} 
We compare to the method from \citet{salesky2019acl} as a strong end-to-end baseline. 
Here, phone \textit{boundaries} are used to segment and compress speech feature vector sequences.
Within each utterance, the feature vectors of consecutive speech frames with the same phone label are averaged to produce one feature vector for translation from a variable number of frames. 
This significantly reduces source sequence lengths (by $\sim$80\%), reducing the number of model parameters and memory. 
Rather than having a variable number of feature vectors per phone-like unit, each has one representation, more similar in granularity to character-based MT.
The averaged feature vectors remain continuously-valued, and are locally summarized: a given phone across the corpus will still have different representations in each instance.

%---------------------------------------------
\section{Data}
We use the Fisher Spanish-English corpus,\footnote{~\url{joshua.incubator.apache.org/data/fisher-callhome-corpus}} which consists of parallel speech, transcripts, and translations, enabling comparisons between cascaded and direct models on the same data and allowing us to generate phone supervision using matched data. 
The dataset contains 160 hours of Spanish telephone speech, split into 138K utterances, which were translated via crowdsourcing by \citet{post2013improved}.
We use the standard dev and test sets, each with $\sim$4k utterances. 
Because we are particularly interested in how our methods will affect training across differently-resourced conditions, we compare results using randomly selected 40 hour and 20 hour subsets of the data.

%--------------------------------------------
\section{Generating Phone Supervision}

To generate phoneme-level labels for sequences of speech features, we generate frame-level alignments using a trained speech recognizer. 
Specifically, we extract 40-dimensional Mel filterbank features with per-speaker mean and variance normalization using Kaldi \cite{povey2011kaldi}.
We train an HMM/GMM system on the full Fisher Spanish dataset with the Kaldi recipe \cite{povey2011kaldi}, using the Spanish CALLHOME Lexicon (LDC96L16), and compute per-frame phone alignments with the triphone model (tri3a) with LDA+MLLT features. 
This yields 50 phone labels, including silence (\textless sil\textgreater), noise, and laughter. 

Producing phone alignments uses supervision from a transcript, which inherently does not exist at inference time. 
While phones can be extracted from Kaldi lattices at inference time, we found that our HMM/GMM model was not our best performing ASR model on this dataset -- by greater than 10 WER. 
To leverage our better-performing neural ASR models for phone generation, we create essentially a `2-pass' alignment procedure: first, generating a transcript, and second, using this transcript to force align phones. 
\myref[Table]{alis} shows the mapping between phone quality and the ASR models used for phone feature generation. 
%----------
\begin{table}[!h]
\centering
\resizebox{\columnwidth}{!}{
\begin{tabular}{ccl} \toprule
\textbf{Alignment Quality} & \textbf{WER} & \textbf{ASR Supervision} \\
\midrule
\multicolumn{1}{g}{Gold} & -- & \textit{Gold transcript} \\
\multicolumn{1}{h}{High} & 23.2 & \citet{salesky2019acl} \\
\multicolumn{1}{m}{Med}  & 30.4 & Seq2Seq ASR \\
\multicolumn{1}{w}{Low}  & 35.5 & Kaldi HMM/GMM \\
\bottomrule 
\end{tabular}}
\caption{Mapping between phone quality and the ASR models used for alignment generation, with the models' WER on Fisher Spanish test.}
\label{alis}
\end{table}
This procedure enables us to both improve phone alignment quality and also match training and inference procedures for phone generation for our translation models. 
In \myref[Section]{analysis}, we compare the impact of phone alignment quality on our translation models utilizing phone features, and show higher quality phone features can improve downstream results by \textgreater10 BLEU.\looseness=-1

Producing phone features in this way uses the same data (source speech and transcripts) as the ASR task in a cascade, and auxiliary ASR tasks from multi-task end-to-end models, but as we show, to far greater effect. 
Further, auxiliary tasks as used in previous work rely on three-way parallel data, while it is possible to generate effective phoneme-level supervision using a recognizer trained on other corpora or languages \citep{salesky2019acl}, though we do not do this here.\looseness=-1

%---------------------------------------------
\section{Model \& Training Procedure} % |[]
As in previous academic work on this corpus \citep{bansal2018low,sperber2019attention,salesky2019acl}, we use a sequence-to-sequence architecture inspired by \citet{weiss2017sequence} modified to train within lower resources; specifically, each model converges within $\approx$5 days on one GPU. 
We build encoder-decoder models with attention in \texttt{xnmt} \citep{neubig2018xnmt} with 512 hidden units. 
Our pyramidal encoder uses 3-layer BiLSTMs with linear network-in-network (NiN) projections and batch normalization between layers \citep{sperber2019attention,zhang2017very}. 
The NiN projections are used to downsample by a factor of 2 between layers, resulting in the same total $4\times$ downsampling in time as the additional convolutional layers from \citet{weiss2017sequence,bansal2018pre}:
They give us the benefit of added depth with fewer additional parameters. 
We use single layer MLP attention \citep{bahdanau2014neural} with 128 units and 1 decoder layer as opposed to 3 or 4 in previous work -- we did not see consistent benefits from additional depth. 
\looseness=-1

In line with previous work on this dataset, all experiments preprocess target text by lowercasing and removing punctuation aside from apostrophes. 
We use 40-dimensional Mel filterbank features as previous work did not see significant difference with higher-dimensional features \citep{salesky2019acl}. 
We use 1k BPE units for translation text, shown in \citet{salesky2019acl} to have both better performance and training efficiency than characters \citep{weiss2017sequence, sperber2019attention} or words \cite{bansal2018low}. 
For both text and phones, we use 64-dimensional embeddings. 
\looseness=-1

For the MT component in cascaded speech translation models, we compared using the pyramidal speech architecture above (3 encoder, 1 decoder layers) to the traditional BiLSTM text model (2 layers each for encoder and decoder). 
Using the pyramidal architecture resulted in the same performance as the BiLSTM model when translating BPE transcriptions from ASR, but gave us consistent improvements of up to 1.5 BLEU when instead translating phone sequences; 
we posit this is because phone sequences are longer than BPE equivalents. 
Accordingly, we use the same model architecture for all our ASR, MT, and ST models. 
\looseness=-1

We use layer dropout with $p=0.2$ and target embedding dropout with $p=0.1$ \cite{gal2016theoretically}.
We apply label smoothing with $p=0.1$ \cite{szegedy2016rethinking} and fix the target embedding norm to 1 \cite{nguyen2017improving}.
For inference, we use beam of size 15 and length normalization with exponent 1.5.
We set the batch size dynamically depending on the input sequence length with average batch size was 36.
We use Adam \cite{kingma2014adam} with initial learning rate 0.0003, decayed by 0.5 when validation BLEU did not improve for 10 epochs initially and subsequently 5 epochs. 
We do not use L2 weight decay or Gaussian noise, and use a single model replica.
We use input feeding \cite{Luong2015b}, and exclude utterances longer than 1500 frames in training for memory.\looseness=-1

%---------------------------------------------
%%HERE TO FORCE LATEX PLACEMENT ON NEXT PAGE
\begin{table*}[b!]
\centering
\setlength\tabcolsep{7pt} % default value: 6pt
\begin{tabular}{llllccccrr} \toprule
  & & \multicolumn{2}{c}{\textbf{HIGH} (160hr)} & \multicolumn{2}{c}{\textbf{MID} (40hr)} & \multicolumn{2}{c}{\textbf{LOW} (20hr)} & \multicolumn{2}{c}{\textit{Components}} \\ 
\cmidrule(lr){3-4} \cmidrule(lr){5-6} \cmidrule(lr){7-8} 
\bf Model & \bf Source & dev & test  & dev & test   & dev & test   & \textit{ASR\,$\downarrow$} & \textit{MT\,$\uparrow$} \\ 
\cmidrule(lr){1-8} \cmidrule(lr){9-10} 
\multirow{3}{*}{Cascaded} & \citet{weiss2017sequence}    & 45.1 & 45.5  & -- & --  & -- & --   & \it 23.2 &  \it 57.9 \\
  & \citet{kumar2014some} & -- &  \textbf{40.4}$^\dagger$  & -- & --  & -- & --  & \it 25.3 & \it 62.9 \\
  & \citet{sperber2019attention} & --   & 32.5  & -- & 16.8  & -- & ~6.6 & \it 40.9 & \it 58.1 \\
\midrule
\multirow{4}{*}{End-to-End} & \citet{weiss2017sequence} & 46.5 & \textbf{47.3}$^*$  & -- & --  & -- & --   \\
 & \citet{salesky2019acl}       & 37.6 & 38.8 & 21.0 & 19.8 & 11.1 & 10.0   \\
 & \citet{sperber2019attention} & --   & 36.7 & --   & \textbf{31.9} & -- & \textbf{22.8}  \\ %attn-passing
%  & \citet{bansal2018low}        & 29.5 & 29.4  & 13.3 & 13.6 & ~5.1  & ~5.3  \\ 
%  & \citet{sperber2019attention} & -- & 35.3  & -- & 14.9  & -- & 6.1   \\ %this is his baseline
%  & \citet{bansal2018pre}        & --   & --   & --   & --   & 10.8 & 10.8   \\
 & \citet{stoian-bansal-pretraining} & 34.1 & 34.6 & --   & --   & 10.3 & 10.2   \\ %baseline
%
%  & Our Baseline                 & 32.4 & 33.7 & 19.5 & 17.4 & ~9.8 & ~9.8 \\
%
\cmidrule(lr){2-8} 
\multirow{2}{*}{\textit{+ Add'l Data}} & \citet{sperber2019attention} & -- & 38.8 & -- & -- & -- & -- \\
  & \citet{stoian-bansal-pretraining} & 37.9 & 37.8 & -- & --  & 20.1 & 20.2   \\ %both dataaug + pretrain
\bottomrule 
\end{tabular}
\caption{End-to-end vs cascaded speech translation model performance in BLEU$\uparrow$ on Fisher Spanish-English data from the literature. ($\dagger$) denotes the best previous academic result on the full dataset, ($*$) the best from industry. Component models for cascades reported on test on full dataset: ASR reported in WER$\downarrow$ and MT in BLEU$\uparrow$.}
\label{prev work}
\vspace{-1em}
\end{table*}
%----------

%---------------------------------------------
\section{Prior Work: Cascaded vs End-to-End Models on Fisher Spanish-English}
\label{sec:baseline}

The large body of research on the Fisher Spanish-English dataset, including both cascaded and end-to-end models, makes it a good benchmark to compare these architectures. 
Not all previous work has compared across multiple resource settings or compared to cascaded models, which we address in this section. 
We summarize best previous results on this dataset on high, medium, and low-resource conditions in \myref[Table]{prev work}.

\paragraph{Best Results.}
The cascade of traditional HMM/DNN ASR and Joshua MT models from \citet{kumar2014some} set a competitive baseline on the full dataset (40.4 test BLEU) which no subsequent academic models have been able to match until this work; subsequent exploration of end-to-end models has produced notable relative improvements but the best end-to-end academic number \citep{salesky2019acl} remains 1.6 BLEU behind this traditional cascade. 

Industry models from \citet{weiss2017sequence} achieved exceptional performance with very deep end-to-end models on the full dataset (47.3 test BLEU), exceeding a cascade for the first time. 
They additionally show results with an updated cascade using neural models, improving over \citet{kumar2014some}. 
Their results have been previously unmet by the rest of the community. 
This is likely in part due to the computational resources required to fully explore training schedules and hyperparameters with models of their depth. 
While their ASR models took $\sim$4 days to converge, their ST models took another 2 weeks, compared to the lighter-weight models of recent academic work which converged in \textless 5 days \cite{sperber2019attention,salesky2019acl,bansal2018pre}. 

This dataset is challenging: improving ASR WER from 35 (\citeauthor{kumar2014some}) to 23 (\citeauthor{weiss2017sequence}) only resulted in 4 BLEU ST improvement: see \textit{Components} in \myref[Table]{prev work}.
We believe this to be in part because the multi-reference scoring masks some model differences, and the conversational phenomena (like disfluencies) are challenging.

\paragraph{Lower-Resource.}
While deep end-to-end models have become competitive at higher-resource conditions, previous work on this dataset has showed they are not as data-efficient as cascades under lower-resource conditions.  
While some works have tested multiple resource conditions, only \citet{sperber2019attention} compared against cascades across multiple conditions. 
Their end-to-end baseline outperformed their cascades on the full dataset, but not under lower-resource conditions, while their end-to-end but multi-stage attention-passing model is more data-efficient than previous models and shows the best previous results under lower-resource condition. 
\citeauthor{sperber2019attention} do not report results without auxiliary ASR, MT, and autoencoding tasks, which they state add up to 2 BLEU.\looseness=-1

\paragraph{Additional Data.} 
\citet{stoian-bansal-pretraining,bansal2018pre,sperber2019attention} investigate speech translation performance using additional corpora through transfer learning from ASR and auxiliary MT tasks. 
The ability to leverage non-parallel corpora was previously a strength of cascades and had not been explored with end-to-end models. 
%\citet{bansal2018pre,sperber2019attention}  also investigate the capacity of different models given additional external corpora, in the former case by pre-training on 300 hours of English ASR data and the latter, using 61M sentences of Spanish-English text from OpenSubtitles for auxiliary tasks. 
We do not use additional data here, but show these numbers as context for our results with phone supervision, and refer readers to \citeauthor{sperber2019attention} for discussion of cascaded and end-to-end models' capacity to make use of more data.\looseness=-1

%---------------------------------------------
\begin{table*}[b!]
\centering
\setlength\tabcolsep{7pt} % default value: 6pt
\begin{tabular}{llccrccrccr} \toprule
  & & \multicolumn{3}{c}{\textbf{HIGH} (160hr)} & \multicolumn{3}{c}{\textbf{MID} (40hr)} & \multicolumn{3}{c}{\textbf{LOW} (20hr)}  \\ 
\cmidrule(lr){3-5} \cmidrule(lr){6-8} \cmidrule(lr){9-11} 
  & \bf Model & dev & test & $\Delta$ & dev & test & $\Delta$ & dev & test &  $\Delta$ \\ 
\cmidrule(lr){1-11} 
\parbox[t]{2mm}{\multirow{3}{*}{\rotatebox[origin=c]{90}{\textit{Baseline}}}} & Baseline End-to-End   & 32.4 & 33.7 & --~ & 19.5 & 17.4 & --~ & 9.8 & 9.8 & --~~ \\
   & \citet{salesky2019acl}  & 37.6 & 38.8 & \dd{5.2} & 21.0 & 19.8 & \dd{2.0} & 11.1 & 10.0 & \dd{0.8} \\
   & Baseline Cascade & 39.7 & 41.0 & \dd{7.3}  & 29.8 &  27.1 & \dd{10.0} & 22.6 & 20.2 & \dd{11.6} \\
\midrule
\parbox[t]{2mm}{\multirow{3}{*}{\rotatebox[origin=c]{90}{\textit{Proposed}}}}  & Phone End-to-End & 40.5 & 42.1 & \dd{8.3}  & 34.5 & 33.0  & \dd{15.3} & 26.7 & 26.2 & \dd{16.7} \\ 
   & Phone Cascade    & 41.6 & 43.3 & \dd{9.4}  & \textbf{37.2} & \textbf{37.4}  & \dd{18.9} & \textbf{32.2} & \textbf{31.5} & \dd{22.1} \\
   \addlinespace[0.1cm]
   & Hybrid Cascade & \textbf{42.9} & \textbf{45.0} & \dd{10.9}  & 33.3 & 31.2 & \dd{13.8} & 23.2 & 21.5 & \dd{12.6} \\
\bottomrule 
\end{tabular}
\caption{Results in BLEU$\uparrow$ comparing our proposed phone featured models to baselines. We compare three resource conditions, and show average improvement for dev and test ($\Delta$). Best performance bolded by column. 
}
\label{summary}
\end{table*}
%----------

\paragraph{Parameter Tuning.}
We find cascaded model performance can be impacted significantly by model settings such as beam size and choice of ASR target preprocessing. 
While \citet{weiss2017sequence,sperber2019attention} use character targets for ASR, we use BPE, which gave us an average increase of 2 BLEU. 
Further, we note that search space in decoding has significant impact on cascaded model performance. In cascaded models, errors produced by ASR can be unrecoverable, as the MT component has access only to ASR output.
While \citet{sperber2019attention} use a beam of size 1 for the ASR component of their cascade to compare with their two-stage end-to-end models, we find that using equal beam sizes of 15 for both ASR and MT improves cascaded performance with the same model by 4-8 BLEU; combining these two parameter changes makes the same cascaded model a much more competitive baseline (compare line 3 in both \myref[Table]{prev work} and \myref[Table]{summary}). 
In contrast, widening beam size to yield an equivalent search space for end-to-end models has diminishing returns after a certain point; we did not see further benefits with a larger beam ($>15$).\looseness=-1

\paragraph{Our Baselines.} 
\label{ourbaselines}
We report best numbers from previous work in \myref[Table]{prev work} for comparison (which may use multi-task training), but use single-task models in our work. 
We report our baseline results in \myref[Table]{summary}. 
On the full dataset, our baseline cascade improves slightly over \citet{kumar2014some} with 41.0 compared to 40.4 on test, a mark most recent work has not matched primarily due to model choices noted above, with component ASR performance of WER 30.4 and 58.6 BLEU for MT. 
Our end-to-end baseline is comparable to the baselines in \citet{salesky2019acl,sperber2019attention,stoian-bansal-pretraining}. 
This suggests we have competitive baselines for both end-to-end and cascaded models.
\looseness=-1

%---------------------------------------
\section{Results Using Phone Features}

We compare our two ways to leverage phone features to our cascaded and end-to-end baselines across three resource conditions. 
\myref[Table]{summary} shows our results; following previous work, all BLEU scores are multi-reference. 
Average single reference scores may be found in \myref[Appendix]{1ref}. 
All models using phone supervision outperform the end-to-end baseline on all three resource conditions, while our proposed models also exceed the cascaded baseline and previous work at lower-resource conditions. 

\paragraph{Phone features.}
\citet{salesky2019acl} performs most similarly to the end-to-end baseline, but nonetheless represents an average relative improvement of 13\% across the three data sizes with a significant reduction in training time. 
Our phone featured models use not just the phone segmentation, but also the phone labels, and perform significantly better. 
Our \textbf{phone end-to-end} model not only shows less of a decrease in performance across resource conditions than \citet{salesky2019acl}, but further improves by 4 BLEU over the baseline cascade on our two lower-resource conditions. 
This suggests augmenting embeddings with discrete phone features is more effective than improved downsampling. 
The \textbf{phone cascade} performs still better, with marked improvements across all conditions over all other models (see \myref[Figure]{sys compare plot}). 
On the full dataset, using phones as the source for MT in a cascade performs $\sim$2 BLEU better than using BPE, while at 40 and 20 hours this increases to up to 10 BLEU.  
We analyze the robustness of phone models further in \myref[Section]{analysis}.

\begin{figure}[!t]
  \includegraphics[width=0.95\linewidth]{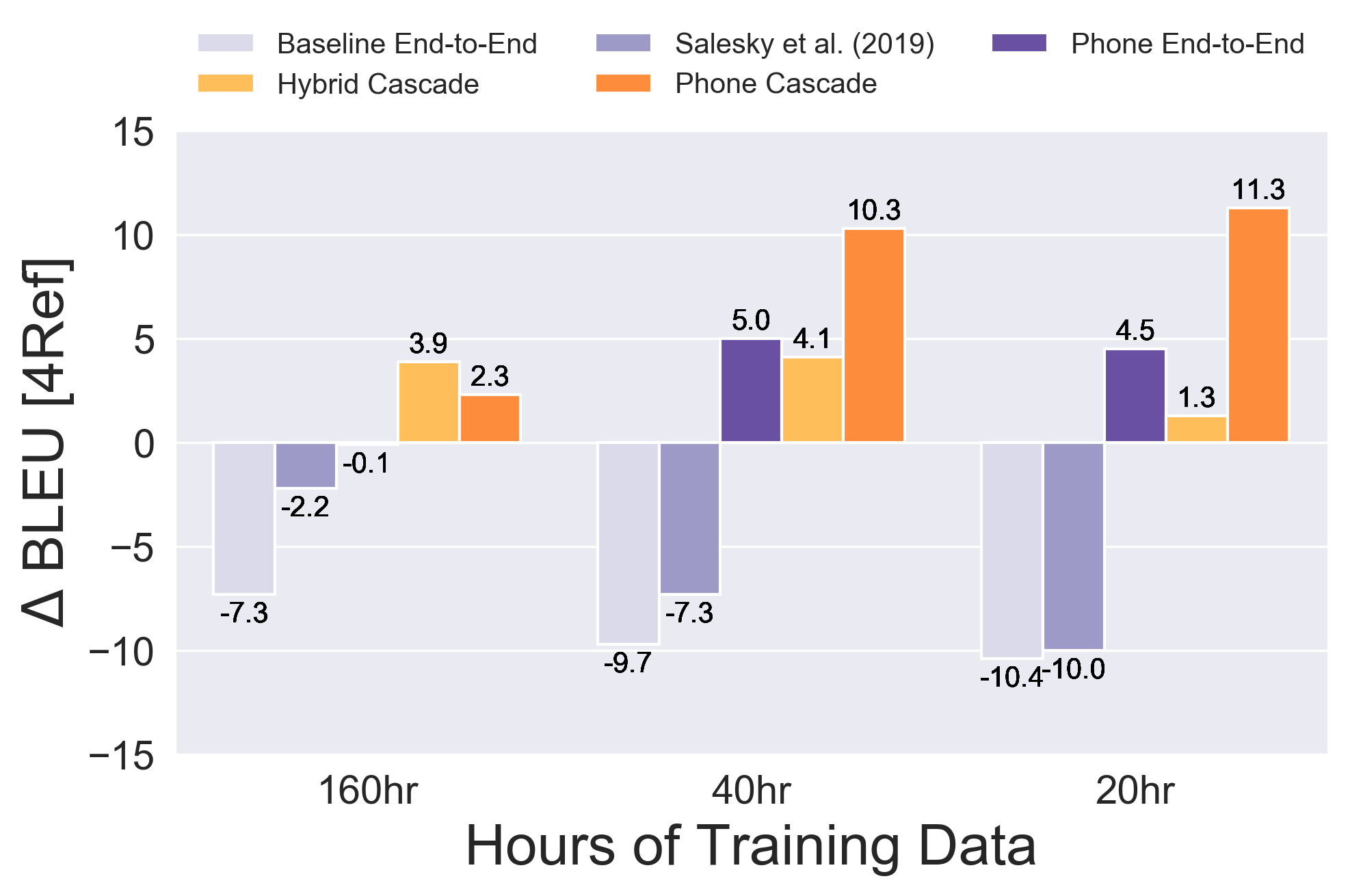}
  \caption{Performance of all models relative to `\textit{Baseline Cascade}' ($\Delta=0$) across our 3 resource conditions. \\
  \textbf{\orange{Cascaded}} models in \orange{\textbf{orange}}, \textbf{\purple{end-to-end}} models in \purple{\textbf{purple}}. Our proposed models yield improvements across all three conditions, with a widening margin under low-resource conditions for the phone cascade.}
  \label{sys compare plot}
  \vspace{-1em}
\end{figure}

\paragraph{Hybrid cascade.}
We additionally use a `hybrid cascade' model to compare using phone features to improving ASR. 
Our hybrid cascade uses an ASR model with phone-informed downsampling and BPE targets \citep{salesky2019acl}. 
This improves the WER of our ASR model to 28.1 on dev and 23.2 on test, matching \citet{weiss2017sequence}'s state-of-the-art on test (23.2) and approaching it on dev (25.7). 
Our hybrid cascade performs more similarly to \citeauthor{weiss2017sequence}'s  cascade on the full dataset, with 45.0 to their 45.5 on test, and is our best-performing ST model on the full dataset. 
However, at lower-resource conditions, it does not perform as favorably compared to phone featured models -- 
as shown in \myref[Figure]{sys compare plot}, both the phone cascade and phone end-to-end models outperform the hybrid cascade at lower-resource conditions, by up to 10 BLEU at 20 hours. 
This suggests improving ASR may enable cascades to perform better at high-resource conditions, but under lower-resource conditions it is not as effective as utilizing phone features. 

\paragraph{Training time.} 
In addition to performance improvements, our models with phone features are typically more efficient with respect to training time, shown in \myref[Table]{times}. 
The fixed time to produce phone labels, which must be performed before translation, becomes a greater proportion of overall training time at lower-resource settings. 
In particular, the phone end-to-end model offers similar training time reduction over the baseline to \citet{salesky2019acl}, where downsampling reduces sequence lengths by up to 60\%, with unreduced sequence lengths through earlier convergence; this model offers a better trade-off between time and performance.

%----------
\begin{table}[!h]
\centering
\setlength\tabcolsep{4pt} % default value: 6pt
\begin{tabular}{lcccc} \toprule
\textbf{Model} & \textbf{HIGH} & \textbf{MID} & \textbf{LOW} & $\Delta$ \\ 
\midrule
Baseline End-to-End    & 118hr & 40hr & 22hr & -- \\
\citet{salesky2019acl} & ~41hr & 13hr & 10hr & \cld{0.4} \\
Baseline Cascade       & ~76hr & 19hr & 12hr & \cld{0.6} \\
\midrule
Phone Cascade    & ~57hr & 39hr & 27hr & \cld{0.7} \\
Phone End-to-End & ~42hr & 20hr & 13hr & \cld{0.4} \\ 
Hybrid Cascade     & ~47hr & 34hr & 24hr & \cld{0.6} \\
\bottomrule 
\end{tabular}
\caption{Total training time \clk ~for all models (including time to generate phone features) on 3 resource conditions. The ASR and MT models in the baseline cascade can be trained in parallel, reflected here, while phone featured models may not as the MT requires phone features from ASR.}
\label{times}
\end{table}
%----------

\paragraph{Comparing to previous work using additional data. }
Previous work used the parallel speech transcripts in this dataset for auxiliary tasks with gains of up to 2 BLEU; we show using the same data to generate phone supervision is far more effective. 
We note that our phone models further outperform previous work trained with additional corpora. 
The attention-passing model of \citet{sperber2019attention} trained on additional parallel Spanish-English text yields 38.8 on test on the full dataset, which  \citet{salesky2019acl} matches on the full dataset and our proposed models exceed, with the phone cascade yielding a similar result (37.4) trained on only 40 hours. 
Pre-training with 300 hours of English ASR data and fine-tuning on 20 hours of Spanish-English data, \citet{stoian-bansal-pretraining,bansal2018pre} improve their end-to-end models from $\approx$10 BLEU to 20.2. 
All three of our proposed models exceed this mark trained on 20 hours of Fisher. 
%----------------------------------------------
\section{Model Robustness \& Further Analysis}
\label{analysis}

In this section, we analyze the robustness of each of our models by varying the quality of our phone features, and further explore the strengths and limitations of each model. 

\subsection{Phone Cascade}

Phone cascades use a representation for translation which may be more robust to non-phonetic aspects of orthography. 
However, as a cascaded model, this still requires hard decisions between ASR and MT, and so we may expect lower phone quality to lead to unrecoverable errors. 
\myref[Figure]{phncascades} compares the impact of phone quality on the performance of phone cascades trained on our high, medium, and low-resource conditions. 
We use alignments produced with gold transcripts as an upper bound on performance. 
We note that with gold alignments, translation performance is similar to text-based translation (see \myref[Section]{ourbaselines}).
We see that phone quality does have a significant impact on performance, with the MT model trained on low phone quality yielding similar translation performance using the full 160 hour dataset to the MT model with the highest quality phones trained on only 20 hours. 
However, we also see significantly more data-efficiency with this model, with less reduction in performance between $160hr\rightarrow40hr\rightarrow20hr$ training conditions than previous models. 
%----------
\begin{figure}[h]
% \vspace{-1em}
\centering
  \includegraphics[width=1\linewidth]{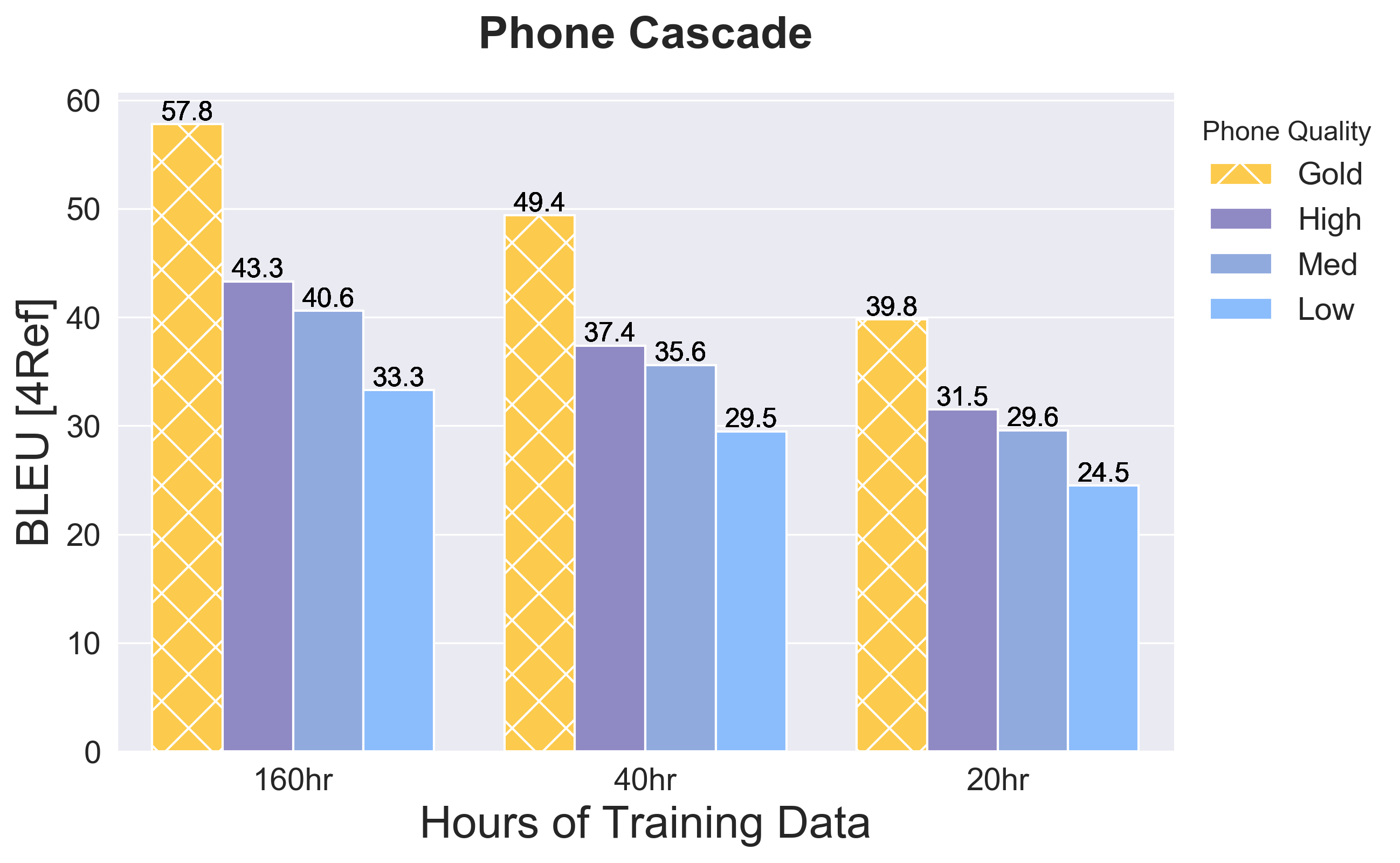}
  \caption{\textbf{Phone Cascade Robustness}: using phone labels in place of BPE as the text source for downstream MT. Comparing performance across our three data conditions and phone label qualities.}
  \label{phncascades}
\end{figure}
%----------
\paragraph{Redundancy.} 
For the phone cascade models compared in \myref[Figure]{phncascades}, we collapse adjacent consecutive phones with the same label, i.e. when three consecutive frames have been aligned to the same phone label \textit{`B B B'} we have reduced the sequence to a single phone \textit{`B'} for translation. 
We additionally compared translating non-uniqued phone sequences (e.g. the same sequence length as the number of frames) as a more controlled proxy for our model's handling of longer frame-based feature vector sequences compared to \citet{salesky2019acl}'s downsampled feature vector sequences. 
The redundant phones caused consistent decreases in BLEU, with much greater impact in lower-resource conditions. 
Translating the full sequence of redundant frame-level phone labels, for the full 160hr dataset, all models performed on average 0.6 BLEU worse; for 40hr, 1.8 BLEU worse; and with 20 hours, 4.1 BLEU worse -- a 13\% decrease in performance \textit{solely from non-uniqued sequences}. 

Phones correspond to a variable-length number of speech frames depending on context, speaker, and other semantic information. 
When translating speech feature vectors, speech features within a phone are similar but uniquely valued; using instead phone labels in a phone cascade, the labels are identical though still redundant. 
These results suggest our LSTM-based models are better able to handle redundancy and variable phone length at higher resource conditions with sufficient examples, but are less able to handle redundancy with less training data.

\subsection{Phone End-to-End}

Our phone end-to-end model concatenates trainable embeddings for phone labels to frame-level filterbank features, associating similar feature vectors \textit{globally} across the corpus, as opposed to \textit{locally} within an utterance with the phone-averaged embeddings. 
\myref[Figure]{phnfactor} compares the results of these factored models using phone features of differing qualities, with `gold' alignments as an upper bound. 
The phone end-to-end models compared do not reach the same upper performance as the phone cascades: comparing gold phone labels, the phone end-to-end model performs slightly worse at 160hr with more degradation in performance at 40hr and 20hr. 
While this comparison is even more pronounced for `low' phone quality than `gold,' the phone end-to-end model has more similar performance between `gold' and `high' phone quality than the cascade. 

%----------
\begin{figure}[!b]
% \vspace{-1em}
\centering
  \includegraphics[width=1\linewidth]{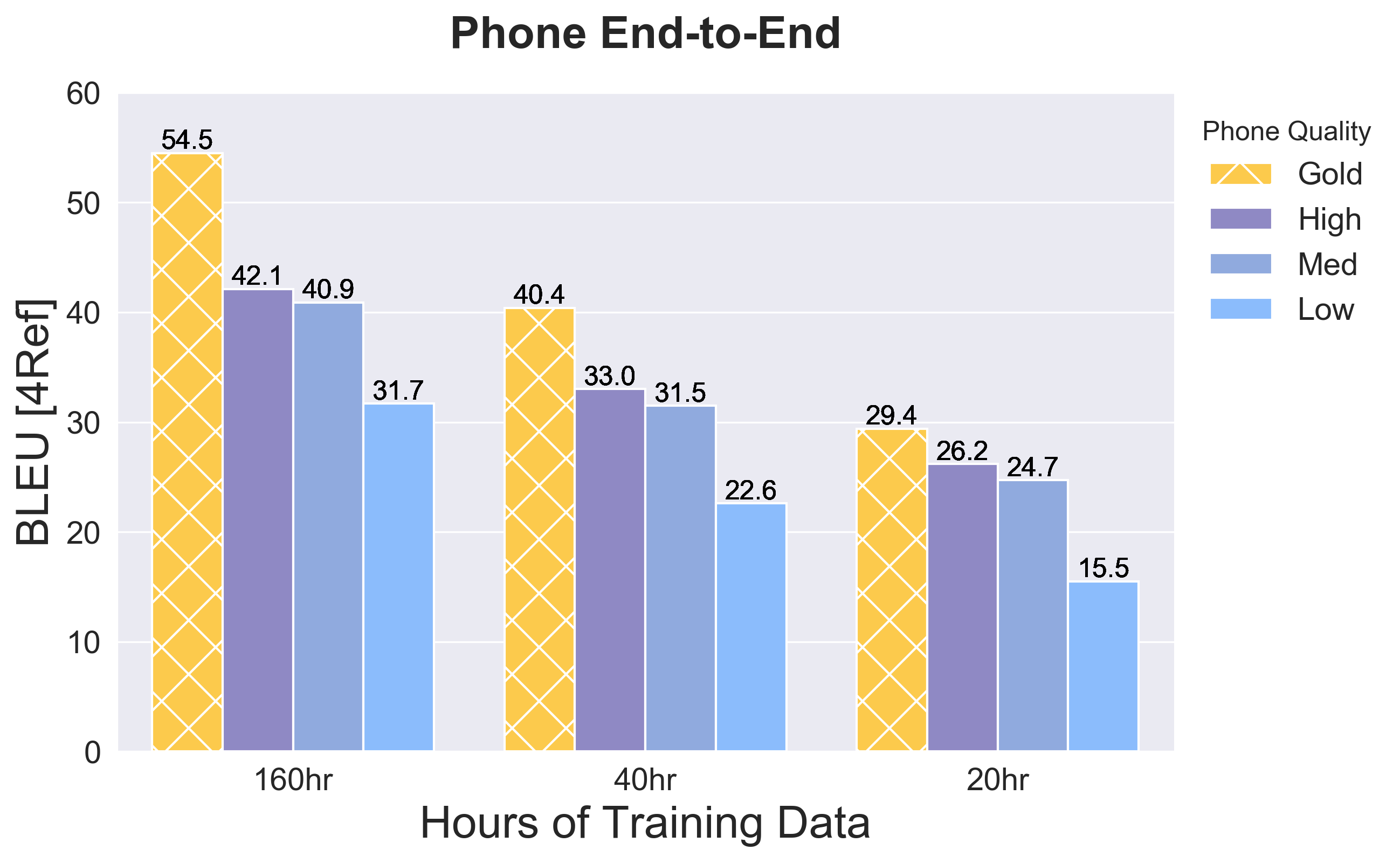}
  \caption{\textbf{Phone End-to-End Robustness}: trainable embeddings for phone labels are concatenated to frame-level filterbank features. Comparing performance across three data conditions and phone label qualities.}
  \label{phnfactor}
\end{figure}
%----------

This model's input contains both the phone features used in the phone cascade and speech features of the baseline end-to-end model, but unlike the phone cascade or \citet{salesky2019acl} the input sequence has not been reduced in length. 
That the end-to-end phone model achieves top performance and converges much faster than end-to-end baseline is unsurprising, as access to both speech feature vectors and phone labels mitigates the effects of long noisy input sequences. 
The significant performance improvements over \citet{salesky2019acl}, however, are more interesting, as these models make use of the similar information in different ways -- the use of discrete embeddings seems to aid the phone end-to-end model, though the sequence length is not reduced. 
The model's performance degradation compared to the phone cascade in lower-resource conditions is likely due in part to these sequence lengths, as shown by our additional experiments with input redundancy for the cascade. 
The greater reduction in performance here using lower quality phones suggests the noise of the labels and concatenated filterbank features compound, further detracting from performance. 
Perhaps further investigation into the relative weights placed on the two embedding factors over the training process could close this additional gap.

\subsection{Phone Segmentation: \citet{salesky2019acl}}

We also compare to the models from \citet{salesky2019acl} as a strong end-to-end baseline. 
That work introduced downsampling informed by phone segmentation -- unlike our other models, the \textit{value} of the phone label is not used, but rather, phone alignments are used only to determine the \textit{boundary} between adjacent phones for variable-length downsampling. 
Their model provides considerable training and decoding time improvements due to the reduced source sequence length, and shows consistent improvements over the baseline end-to-end model using the original filterbank feature sequences which increase with the amount of training data. 
However, their model has lower overall performance and with much smaller performance improvements over our baselines in lower-resource conditions than the phone featured models we propose here. 
We hypothesize that the primary reason for their BLEU improvements is the reduction in local redundancy between similar frames, as discovered in the previous section. 
We refer readers to their paper for further analysis.

\subsection{Quality of Phone Labels}

We show two examples of phone sequences produced with each overall model quality in \myref[Figure]{phn examples}, uniqued within consecutive frame sequences with the same label for space constraints. 
Individual phones are typically 5-20 frames. 
We see the primary difference in produced phones between different models is the label values, rather than the boundaries. 
While we do see some cases where the boundaries shift, they chiefly vary by only 1-3 frames. 
It is not the case that there are significantly more or fewer phone segments aligned per utterance by quality, though there are outlying utterances (Example 2 -- `Low'). 

%----------
\begin{figure}[th]
\centering
  \includegraphics[width=1\linewidth]{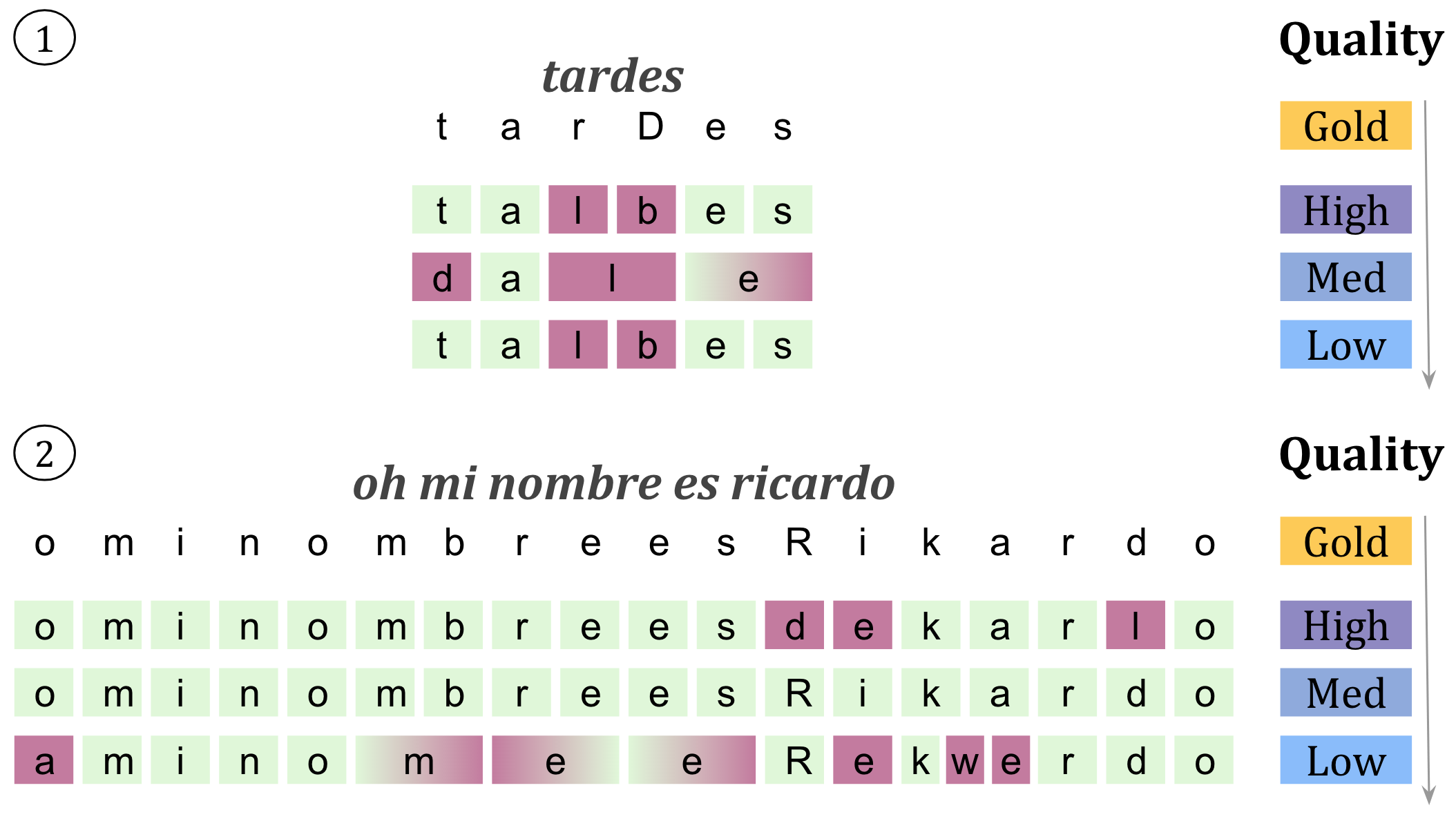}
  \caption{Two examples of phone sequences demonstrating  differences across qualities of phone features. \\
  (See \myref[Table]{alis} for the mapping between quality and generation procedure). Note: word-level segmentation is not marked, as it is also not present in \{\textit{speech,phone}\} source sequences for translation.}
  \label{phn examples}
\end{figure}
%----------

\newpage
Relating our observed trends to the differences between our phone cascades and phone end-to-end models, we note that differences in frame-level phone boundaries would not affect our phone cascaded models, where the speech features are discarded, while they would affect our phone end-to-end models, where the phone labels are concatenated to speech feature vectors and associate them across the corpus. 
While errors in phone labels may be seen as `unrecoverable' in a cascade, for the end-to-end model, they add noise to distribution of filterbank feature associated with each phone label embedding, which appears to have a more negative impact on performance than the hard decisions in cascades. 
Though the concatenated filterbank features may allow our end-to-end models to recover from discrete label errors, our results testing various phone qualities suggest this may only be the case under higher-resource settings with sufficient examples.

%---------------------------------------------

\section{Related Work}

Speech translation was initially performed by cascading separately trained ASR and MT models, allowing each model to be trained on larger data sources without parallel speech, transcriptions, and translations, but potentially yielding unrecoverable errors between models.  
Linking models through lattices with both phrase-based \citep{kumar2014some} and neural MT \citep{sperber2017neural} reduced many such errors. 
Using one model to directly translate speech was later enabled by attentional encoder-decoder models.

Direct end-to-end speech translation was first explored as a way to reduce both error propagation, and also the need for high quality intermediate transcriptions (e.g. for unwritten languages). 
The first such models were investigated in \citet{berard2016listen,duong2016attentional}, but these used, respectively, a small synthetic corpus and evaluated on speech-to-text alignments rather than translation. 
Subsequently \citet{weiss2017sequence} extended these neural attentional models to deep, multi-task models with excellent results on Fisher Spanish--English, exceeding a cascade for the first time. 
However, efforts from the community have not yet replicated their success \cite{stoian-bansal-pretraining,sperber2019attention,salesky2019acl}. 
End-to-end models have performed inconsistently compared to cascades on other corpora: 
\citet{berard2018end} perform well on high-resource audiobooks but do not exceed a cascade; 
\citet{anastasopoulos2018tied} found `triangle' models performed better than cascades for 2 of 3 very low-resource language pairs; 
and in the most recent IWSLT evaluation campaigns, cascades have remained the highest-performing systems \citep{niehues2018iwslt,niehues2019iwslt}.

Similarly-motivated work exists in speech translation. 
In addition to \citet{salesky2019acl,sperber2019attention} addressed above, preliminary cascades using phone-like units have been explored for low-resource speech translation, motivated by translation of unwritten languages where a traditional cascade would not be possible.  
To this end, \citet{bansal2018low} utilized unsupervised term discovery, and \citet{wilkinson2016subword} synthesized speech; but these approaches were only evaluated in terms of precision and recall and were not tested on both `higher-resource' and natural speech data conditions.

%---------------------------------------------
\section{Conclusion}

We show that phone features significantly improve the performance and data efficiency of neural speech translation models. 
We study the existing performance gap between cascaded and end-to-end models, and introduce two methods to use phoneme-level features in both architectures. 
Our improvements hold across high, medium, and low-resource conditions. 
Our greatest improvements are seen in our lowest-resource settings \textit{(20 hours)}, where our end-to-end model outperforms a strong baseline cascade by $\approx$5 BLEU, and our cascade outperforms prior work by $\approx$9 BLEU. 
Generating phone features uses the same data as auxiliary speech recognition tasks from prior work; our experiments suggest these features are a more effective use of this data, with our models matching the performance from previous works' performance without additional training data.  
We hope that these model comparisons and results inform development of more robust end-to-end models, and provide a stronger benchmark for performance on low-resource settings.

%-------------------------------------------------
\section*{Acknowledgments}
The authors thank Andrew Runge, Carlos Aguirre, Carol Edwards, Eleanor Chodroff, Florian's cluster, Huda Khayrallah, Matthew Wiesner, Nikolai Vogler, Rachel Wicks, Ryan Cotterell, and the anonymous reviewers for helpful feedback and resources.

%-------------------------------------------------
\bibliography{bib}
\bibliographystyle{acl_natbib.bst}
%-------------------------------------------------

\appendix

\onecolumn

\vspace{1em}
\section{Single-Reference BLEU Scores}
\label{1ref}

These tables contain the same results as our tables and figures as in the main paper, but show \textit{average single-reference} BLEU scores in place of \textit{multi-reference (4-reference)} BLEU. 
WER for ASR is unchanged: the dataset contains a single reference transcript for ASR. 
Results from prior work report only multi-reference BLEU and so are not included below. 
~\\

%----------
\begin{table*}[h]
\centering
\setlength\tabcolsep{7pt} % default value: 6pt
\begin{tabular}{llccrccrccr} \toprule
 & & \multicolumn{3}{c}{\textbf{Full (160hr)}} & \multicolumn{3}{c}{\textbf{40hr}} & \multicolumn{3}{c}{\textbf{20hr}} \\ 
\cmidrule(lr){3-5} \cmidrule(lr){6-8} \cmidrule(lr){9-11} 
 & \bf Model & dev & test & $\Delta$ & dev & test & $\Delta$ & dev & test & $\Delta$ \\ 
\midrule
\parbox[t]{2mm}{\multirow{3}{*}{\rotatebox[origin=c]{90}{\textit{Baseline}}}} & Baseline End-to-End   & 19.0 & 19.6 & -- & 11.5 & 10.4 & -- & ~5.9 & ~5.3 & -- \\
 & \citet{salesky2019acl} & 22.0 & 21.9 & \dd{2.7} & 12.6 & 11.6 & \dd{1.2} & ~6.7 & ~6.2 & \dd{0.9}  \\
 & Baseline Cascade & 23.2 & 23.7 & \dd{4.2} & 17.4 & 15.7 & \dd{5.6} & 13.2 & 11.8 & \dd{6.9} \\
\midrule
\parbox[t]{2mm}{\multirow{3}{*}{\rotatebox[origin=c]{90}{\textit{Proposed}}}} & Phone End-to-End & 24.0 & 23.7 & \dd{4.6} & 20.8 & 18.4 & \dd{8.7} & 16.5 & 14.6 & \dd{10.0} \\ 
 & Phone Cascade    & 24.1 & 25.1 & \dd{5.3} & \textbf{21.6} & \textbf{21.7} & \dd{10.7} & \textbf{18.9} & \textbf{18.3 }& \dd{13.0}  \\
 & Hybrid Cascade     & \textbf{24.9} & \textbf{25.9 }& \dd{6.1} & 19.6 & 18.2 & \dd{8.0} & 13.6 & 12.6 & \dd{7.5}  \\
\bottomrule 
\end{tabular}
\caption{Results in BLEU$\uparrow$ comparing our proposed phone featured models to baselines. We compare three resource conditions, and show average improvement for dev and test ($\Delta$). Best performance bolded by column. 
}
\label{1ref-summary}
\end{table*}
%----------

%----------
\begin{table}[ht]
\centering
\begin{minipage}{.45\linewidth}
\centering
\resizebox{\columnwidth}{!}{
\begin{tabular}{lcccccc} \toprule
\multirow{2}{*}{\parbox{0.9cm}{Phone \\ Quality}} & \multicolumn{2}{c}{\textbf{160hr}} & \multicolumn{2}{c}{\textbf{40hr}} & \multicolumn{2}{c}{\textbf{20hr}} \\ 
\cmidrule(lr){2-3} \cmidrule(lr){4-5} \cmidrule(lr){6-7} 
 & dev & test & dev & test & dev & test \\ 
\midrule
\multicolumn{1}{g}{Gold} & 33.3 & 33.2 & \multicolumn{1}{|c}{29.3} & \multicolumn{1}{c|}{28.5} & 24.4 & 23.0 \\
\midrule
\multicolumn{1}{h}{High} & 24.1 & 25.1 & \multicolumn{1}{|c}{21.6} & \multicolumn{1}{c|}{21.7} & 18.9 & 18.3 \\
\multicolumn{1}{m}{Med}  & 23.1 & 23.4 & \multicolumn{1}{|c}{20.6} & \multicolumn{1}{c|}{20.7} & 17.6 & 17.2 \\
\multicolumn{1}{w}{Low}  & 18.2 & 19.1 & \multicolumn{1}{|c}{16.4} & \multicolumn{1}{c|}{17.0} & 14.1 & 14.2 \\
\bottomrule 
\end{tabular}}
\caption{\textbf{Phone Cascades}. We use frame-level phone labels as the text source for downstream MT. Comparing method robustness to phone quality and resource conditions.}
\label{1ref-phncascades}

\end{minipage}
\hfill
\begin{minipage}{.45\linewidth}
\centering
\resizebox{\columnwidth}{!}{
\begin{tabular}{lcccccc} \toprule
\multirow{2}{*}{\parbox{0.9cm}{Phone \\ Quality}} & \multicolumn{2}{c}{\textbf{160hr}} & \multicolumn{2}{c}{\textbf{40hr}} & \multicolumn{2}{c}{\textbf{20hr}} \\ 
\cmidrule(lr){2-3} \cmidrule(lr){4-5} \cmidrule(lr){6-7} 
 & dev & test & dev & test & dev & test \\ 
\midrule
\multicolumn{1}{g}{Gold} & 34.1 & 31.3 & \multicolumn{1}{|c}{27.9} & \multicolumn{1}{c|}{23.4} & 20.5 & 17.2 \\
\midrule
\multicolumn{1}{m}{Med}  & 24.0 & 23.7 & \multicolumn{1}{|c}{20.8} & \multicolumn{1}{c|}{18.4} & 16.5 & 14.6 \\
\multicolumn{1}{w}{Low}  & 20.5 & 18.3 & \multicolumn{1}{|c}{17.0} & \multicolumn{1}{c|}{13.0} & 12.2 & ~~8.7 \\
\bottomrule 
\end{tabular}}
\caption{\textbf{Phone End-to-End}. Trainable embeddings for phone labels are concatenated to frame-level filterbank features. Comparing method robustness to phone quality and resource conditions.}
\label{1ref-phnfactor}
\end{minipage}

\end{table}
%----------

%---------------------------------------------
\begin{table}[!th]
%------
% to color asr column in booktabs without weird white gaps need to modify height of cells:
\aboverulesep=1pt
\belowrulesep=1pt
\extrarowheight=\aboverulesep
\addtolength{\extrarowheight}{\belowrulesep}
\aboverulesep=0pt
\belowrulesep=0pt
\captionsetup{width=.45\linewidth}
%-----
\centering
\setlength\tabcolsep{3pt} % default value: 6pt
\resizebox{0.45\linewidth}{!}{
\begin{tabular}{cmmwwcccc} \toprule
& \multicolumn{2}{m}{ASR$\downarrow$} &  \multicolumn{2}{w}{MT$\uparrow$} & \multicolumn{2}{c}{Cascade} & \multicolumn{2}{c}{End-to-End} \\\cmidrule(lr){2-3} \cmidrule(lr){4-5} \cmidrule(lr){6-7} \cmidrule(lr){8-9}
\bf Data & \bf dev & \bf test & \bf dev & \bf test & \bf dev & \bf test & \bf dev & \bf test \\ 
\midrule
\bf Full & 33.3 & 30.4 & 34.5 & 33.6 & 23.2 & 23.7 & 19.0 & 19.6 \\
\bf 40hr & 44.8 & 46.7 & 29.9 & 28.3 & 17.4 & 15.7 & 11.5 & 10.4 \\
\bf 20hr & 56.3 & 59.1 & 22.4 & 22.6 & 13.2 & 11.8 & ~5.9 & ~5.3 \\
\bottomrule 
\end{tabular}}
\caption{\textbf{Baseline results} for end-to-end and cascaded speech translation models, with component ASR and MT model performance for cascades (\blue{blue}). ASR results in WER$\downarrow$ and translation results in BLEU$\uparrow$.}
\label{1ref-baselines}
\end{table}
%----------

%------------------------------------------------------------
% ta ta for now!
%------------------------------------------------------------
\end{document}